\documentclass[conference]{IEEEtran}
\IEEEoverridecommandlockouts
\usepackage{cite}
\usepackage{amsmath,amssymb,amsfonts}
\usepackage{algorithmic}
\usepackage{graphicx}
\usepackage{textcomp}
\usepackage{xcolor}
\usepackage{hyperref} 
\usepackage{bm}
\usepackage{graphicx}
\usepackage{amssymb} 
\usepackage{float}
\usepackage{booktabs}
\usepackage{multirow}
\usepackage{array}
\usepackage{makecell}
\usepackage{tabularx}
\usepackage{subcaption} 
\usepackage{tikz}
\usepackage{xcolor}
\usepackage{listings}
\usepackage[affil-it]{authblk}
\usepackage{adjustbox}

\usepackage[flushmargin]{footmisc} %
\makeatletter
\renewcommand{\footnoterule}{%
  \kern-3pt %
  \hrule width 0.5\columnwidth height 0.5pt
  \kern 2.5pt 
}
\makeatother

\definecolor{mypaleblue}{RGB}{10,110,255} %

\def\BibTeX{{\rm B\kern-.05em{\sc i\kern-.025em b}\kern-.08em
    T\kern-.1667em\lower.7ex\hbox{E}\kern-.125emX}}
\begin{document}

\title{A-MESS: Anchor-based Multimodal Embedding with Semantic Synchronization for Multimodal Intent Recognition}


\author{
     Yaomin Shen$^{1}$\textsuperscript{*}\thanks{*Corresponding author, project leader.\\Nanchang Research Institute of Zhejiang University, now also known as
 Jiangxi Qiushi Higher Research Institute. The director of the XR System Ap
plication Research Center is Wenhuai Yu: ricky@caece.net, Taiwan,
 China.}, 
     Xiaojian Lin$^{2}$, 
     Wei Fan$^{3}$, \\
     \emph{$^{1}$XR System Application Research Center, Nanchang Research Institute, Zhejiang University, China} \\ 
     \emph{$^{2}$Institute for AI Industry Research(AIR), Tsinghua University, China} \\
     \emph{$^{3}$Independent Researcher, China} \\
     {\small \tt $^{1}$coolshennf@gmail.com, $^{2}$chrislim@connect.hku.hk, $^{3}$caucfw@gmail.com}
}

\maketitle

\begin{abstract}
In the domain of multimodal intent recognition (MIR), the objective is to recognize human intent by integrating a variety of modalities, such as language text, body gestures, and tones. However, existing approaches face difficulties adequately capturing the intrinsic connections between the modalities and overlooking the corresponding semantic representations of intent. To address these limitations, we present the Anchor-based Multimodal Embedding with Semantic Synchronization (A-MESS) framework. We first design an Anchor-based Multimodal Embedding (A-ME) module that employs an anchor-based embedding fusion mechanism to integrate multimodal inputs. Furthermore, we develop a Semantic Synchronization (SS) strategy with the Triplet Contrastive Learning pipeline, which optimizes the process by synchronizing multimodal representation with label descriptions produced by the large language model. Comprehensive experiments indicate that our A-MESS achieves state-of-the-art and provides substantial insight into multimodal representation and downstream tasks.
\end{abstract}

\section{Introduction}
In the field of natural language understanding, the multimodal intent recognition (MIR) task, used to categorize intent within goal-driven context based on textual, visual and auditory information, has been identified as a critical element in identifying complex human behavioral intent \cite{wu2020tod}. Especially in AI Agent\cite{durante2024agentaisurveyinghorizons} applications, for example, when users need to command the AI agent to do specific tasks, the AI agent can perform the tasks well only if it correctly understands the intent behind the user's commands. Compared to the method \cite{devlin2019bertpretrainingdeepbidirectional} that relies solely on a single data type, the use of multiple data types provides a more substantial information base, which can improve the accuracy of identifying complex intent categories. In this domain, pioneering studies \cite{zhang2022mintrec,zhang2024mintrec} have collected multimodal data from real world settings for the creation of intent recognition datasets, making a significant contribution to MIR research. Current methods \cite{sun2024contextual,zhou2024token,rahman2020integrating,tsai2019multimodal}, excel in the MIR task, however, numerous untapped representational strategies remain to be discovered, such as semantic correlation between labels and multimodal embeddings, and the multimodal information redundancy elimination strategy, which also pose significant challenges. We summarize them as two key challenges in MIR task: \textbf{Challenge. I:} As a text-centered task, utilizing audio and visual modalities as auxiliary signals, most components of these signals could introduce interference to the representation, so finding a strategy to filter out disruptive information and retaining the critical components of the representation becomes the first challenge. \textbf{Challenge. II:} Development of a more efficient learning strategy to optimize the entire MIR algorithmic structure and enhance the joint representation from the aggregation of three modalities.\par
\begin{figure}[!t]
    \centering
     \includegraphics[width=1\columnwidth]{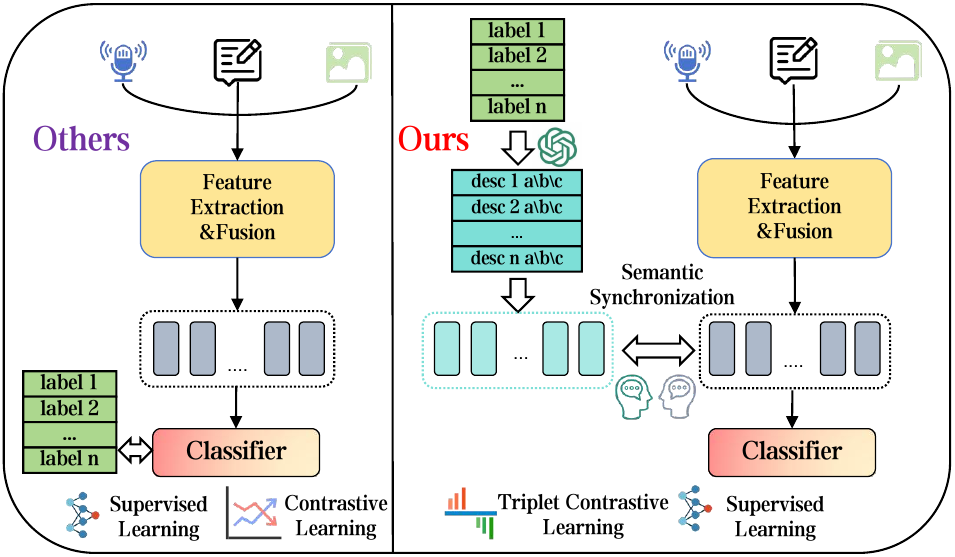}
    \caption{The architecture of the \textbf{(\textit{A-MESS})} framework. Illustration of the \textcolor{violet}{others} (left) frameworks employed in the majority of preceding studies, and \textcolor{red}{ours} (right) performs synchronization of the description generated by the Large Language Model.}
    \label{fig:contra}
\end{figure}
In the \textbf{Challenge. I}, we try to identify key components, which we term ``anchors", from both auxiliary signals and text signals, while filtering out irrelevant information. By fusing and interacting these ``anchors" to effectively address this challenge. To address \textbf{Challenge. II}, we try to synchronize multimodal representations with their intent-oriented semantic information. We think that this can align these representations into a more reasonable semantic space and better optimize the entire learning process.\par

\begin{figure*}[!t]
    \centering
    \includegraphics[width=\textwidth]{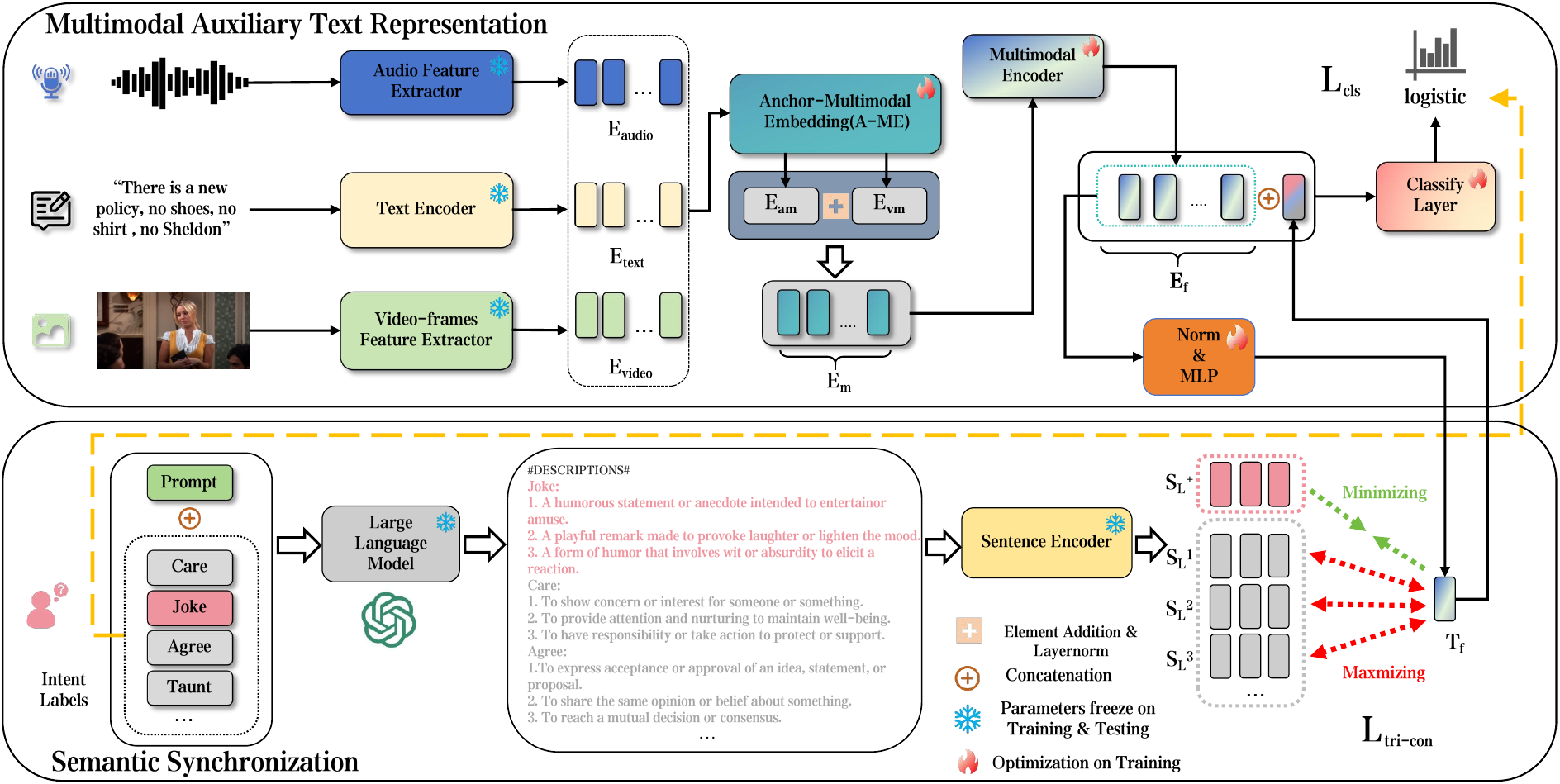} 
    \caption{Overview of the \textbf{(\textit{A-MESS})} architecture, within the Multimodal Auxiliary Text Representation component (top), we initially feed the feature embeddings of audio, images, text into the \textbf{(\textit{A-ME})} module for auxiliary enhancing text embedding. The generated embeddings are then concatenate and feed into multimodal encoder to achieve multimodal integration. In the Semantic Synchronization phase (bottom), we encode multiple description sentences generated by LLM into embeddings, then through triplet contrastive learning with the previously obtained multimodal embeddings. Finally, these embeddings are fed back into the Multimodal Auxiliary Text Representation component for classification computation.}
    \label{fig:framework}
\end{figure*}
\label{sec:intro}

Driven by the above motivation, in this paper, we introduce a new framework: \textbf{\textit{A}}nchor-based \textbf{\textit{M}}ultimodal \textbf{\textit{E}}mbedding with \textbf{\textit{S}}emantic \textbf{\textit{S}}ynchronization \textbf{(\textit{A-MESS})}, as illustrated in Fig. \ref{fig:contra}, which leverages the joint anchors embedding representation derived from audio and video modalities to enhance textual representation. The enhanced textual embeddings are then further integrated into a multimodal encoder. The encoded representation is then synchronized with multiple explanations generated by a Large Language Model (LLM) \cite{achiam2023gpt} from prompted labels using triplet contrastive loss. Simultaneously, the process is jointly optimized by classification loss. Extensive experiments are conducted on the MIntRec\cite{zhang2022mintrec} and MIntRec2.0\cite{zhang2024mintrec} datasets, demonstrating that our approach achieves significant improvements over state-of-the-art methods. Our contributions can be summarized as follows:
\begin{itemize}
    \item We design a Anchor-based Multimodal Embedding \textbf{(\textit{A-ME})} module that selectively generates auxiliary embeddings by integrating video and audio anchors embeddings to assist in the representation of the textual modality.
    \item We develop a Semantic Synchronization \textbf{(\textit{SS})} framework combined with triplet contrastive learning, which utilizes semantic embeddings with positive and negative sample labels generated by LLM for semantic synchronization, and is optimized through the proposed triplet contrastive loss. To the best of our knowledge, we are the first to introduce the LLM representation capability into the MIR learning task.
    \item We conduct comprehensive experiments on two challenging datasets, demonstrating that the proposed method achieves state-of-the-art performance on the MIR task.
\end{itemize}
\section{Related Works}
\subsection{Multimodal Intent Recognition}
Intuitively, the MIR task represents a multimodal fusion task that requires the achievement of high-quality representations in different modalities \cite{ezzameli2023emotion}. \cite{rahman2020integrating} propose MAG-BERT which incorporates an additional mechanism aimed at enhancing the fine-tuning process of BERT \cite{devlin2019bertpretrainingdeepbidirectional}. \cite{zhou2024token} introduced a modality-aware prompt framework that employs token-level contrastive learning, and this framework is designed to effectively integrate multimodal features and adaptively learn prompts across different modalities in the context of the MIR task. \cite{sun2024contextual} improved the expressiveness of the features by extracting contextual interactions within the video frames.
\subsection{Multimodal Embedding Learning}
Embedding learning, as part of representation learning, entails transforming data into fixed-dimensional vectors \cite{zhang2020multimodal,xu2023multimodal}, which are designed to encapsulate the essential characteristics of the original data while maintaining the relationships inherent in the original data. \cite{le2023multi} introduced a Transformer-based fusion and embedding representation learning method to integrate and enrich multimodal features from raw videos to complete the multilabel video sentiment recognition task. In contrast to previous approaches, we design an anchor-based multimodal embedding approach to extract key components from multimodalities for better representation.

\subsection{Contrastive Learning}
Contrastive learning is characterized by its focus on increasing the similarity of close data pairs while enhancing the difference between distant ones. The foundational work \cite{radford2021learningtransferablevisualmodels,chen2020simpleframeworkcontrastivelearning,zhang2022dino} laid the groundwork for the development of contrastive learning. In our framework, a novel triplet contrastive learning approach is utilized to learn rich semantic synchronization. Furthermore, within the proposed framework, optimization is not only achieved through the loss between model output and one-hot labels, but is also constrained by incorporating supplementary descriptions of these labels provided by the LLM (this leveraging robust representation capability of LLM). The entire semantic synchronization training process is fully automated.

\section{Method}
\subsection{Overview}
In this section, we describe the architecture of our proposed \textbf{(\textit{A-MESS})}. As shown in Fig. \ref{fig:framework}: it primarily consists of two major components, the Multimodal Auxiliary Text Representation (top) and Semantic Synchronization (bottom). The former is composed of Multimodal Feature Extraction, Anchor-based Multimodal Embedding, Multimodal Encoder, and Classify layer, executed sequentially. The latter consists of Large Language Model (LLM) Prompting Inference, Embedding of Semantic Interpretive information on the labels, and Semantic Synchronization Triplet Contrastive Learning. 
\subsection{Multimodal Auxiliary Text Representation}
\textbf{Feature Extraction.} Following \cite{zhang2024mintrec,zhou2024token}, we use the embedding layer of a pre-trained BERT \cite{devlin2019bertpretrainingdeepbidirectional} to extract the textual embedding. Specifically, given an input of text $t$, we obtain the embeddings from the BERTEncoder layer:
\begin{equation}
\bm{E}_{text} = \text{BERTEncoder}(t),
\end{equation}
where $\bm{E}_{text} = [CLS_0,e_{l_1},...,e_{l_{t-1}}] \in \mathbb{R}^{{l_t}\times d_t}$ denotes the embedding set, $l_t$ is the length of the text sequence, and $d_t$ is the embedding dimension. \par

We extract embedding from video frames using a pre-trained Swin-Transformer same as \cite{zhou2024token}. Specifically, each segmented video frame $[v_1,v_2,...,v_{l_v}]$ is input into the feature extraction method to obtain the embedding of the last layer:
\begin{equation}
\bm{E}_{video} = \text{Swin-Transformer}([v_1,v_2,...,v_{l_v}]),
\end{equation}
where $\bm{E}_{video} = [e_{l_1},e_{l_2},...,e_{{l_v}}] \in \mathbb{R}^{l_v \times d_v}$ are the embeddings of the video frames, $l_v$ is the number of video frames,  and $d_v$ is the dimension of the video embeddings.\par
To extract audio embeddings, we use a pre-trained WavLM \cite{chen2022wavlm} model, each audio segment $au$ will be processed as:
\begin{equation}
\bm{E}_{audio} = \text{WavLM}(au),
\end{equation}
where $\bm{E}_{audio} = [a_1,a_2,...,a_{l_{a}}] \in \mathbb{R}^{l_a \times d_a}$ denotes the audio embedding from last hidden layer, $l_a$ is the sequence length and $d_a$ is the audio embedding dimension.\par

\begin{figure}[!t]
    \centering
     \includegraphics[width=1\columnwidth]{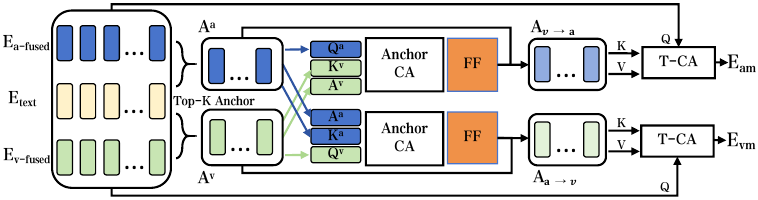} 
    \caption{Architecture of \textbf{(\textit{A-ME})} module. Firstly algin sequence length and dimensions, and then select anchors with top-k ratio. Feed the anchors into the anchor cross attention and feedforward(FF). Finally, the anchor-enhanced embeddings are fed into the temporal cross attention to obtain the final results.}
    \label{fig:EM}
\end{figure}
\noindent\textbf{Anchor-based Multimodal Embedding.} Inspired by \cite{lowe2004distinctive,ZHANG2025107399}, we propose an Anchor-based Multimodal Embedding module (A-ME), as shown in Fig. \ref{fig:EM}. Given $\bm{E}_{video}$, $\bm{E}_{audio}$, $\bm{E}_{text}$, we first perform sequence alignment employing the same method as \cite{rahman2020integrating}. Then we put the aligned embeddings in a cross-attention block to capture the primary modal information and achieve dimension alignment with $\bm{E}_{text}$. The $\bm{E}_{text}$ serves as query and $\bm{E}_{video}$ or $\bm{E}_{audio}$ serves as value and key. Then we get $\bm{E}_{v\text{-}{fused}},\bm{E}_{a\text{-}{fused}} \in \mathbb{R}^{l_t \times d_t}$.\par
To obtain the representative anchor, we rank the token cosine similarity between \{  $\bm{E}_{text}$ and $\bm{E}_{v\text{-}{fused}}$ \} as well as \{$\bm{E}_{text}$ and $\bm{E}_{a\text{-}{fused}}$\}. We compute the ratio between the first and second nearest-neighbor tokens, respectively, which is utilized as a reliability score. Subsequently, the top-k tokens in $\bm{E}_{v\text{-}{fused}}$ and $\bm{E}_{a\text{-}{fused}}$ are selected by score as anchor tokens, we get $\bm{A}^{v} = [an_1^v,an_2^v,...,an_k^v] \in \mathbb{R}^{k \times d_t}$ and $\bm{A}^a = [an_1^a,an_2^a,...,an_k^a] \in \mathbb{R}^{k \times d_t}$, where $k$ denotes the anchor number, the process described above can be formally expressed as:
\begin{equation}
\bm{E}_{v\text{-}{fused}} = Cross\text{-}Attn(Align(\bm{E}_{video}),\bm{E}_{text}),
\end{equation}
\begin{equation}
\bm{A}^{v} = AnSelect(\bm{E}_{text}, \bm{E}_{v\text{-}{fused}}),
\end{equation}
and
\begin{equation}
\bm{E}_{a\text{-}{fused}} = Cross\text{-}Attn(Align(\bm{E}_{audio}),\bm{E}_{text}),
\end{equation}
\begin{equation}
\bm{A}^a = AnSelect(\bm{E}_{text}, \bm{E}_{a\text{-}{fused}}),
\end{equation}

After obtaining the anchors, we apply cross-attention to these two streams to capture the interactions among anchors within each modality, thereby enhancing interactive learning. To be specific, to achieve the propagation of multimodel information from one to another, we project $\bm{A}^{v}$ , $\bm{A}^a$ and $\bm{A}^a$ in query $\bm{Q}^v$, $\bm{K}^a$ and $\bm{V}^a$, respectively, by using three different linear projections. Then, the Anchor Cross-Attention($An\text{-}CAttn$) is employed for these modalities:
\begin{equation}
An\text{-}CAttn_{a \to v}(\bm{Q}^v,\bm{K}^{a}) = Softmax(\frac{{\bm{Q}^v \bm{K}^{a}}}{\sqrt{d_t}}),
\end{equation}
where subscript $a \to v$ denotes vision to audio anchor. Using intensive interaction, the vision anchor embedding $\bm{A}^{v}$ can be further enhanced by adding the information from the audio anchor embedding within its origin anchors as:
\begin{equation}
\bm{A}_{a \to v} = \bm{A}^{v} + FF(An\text{-}CAttn_{a \to v}(\bm{Q}^v,\bm{K}^a)\bm{A}^a),
\end{equation}
where $FF$ denotes Feedforward \cite{vaswani2017attention}. Vice versa, we can compute the embedding of the audio to vision anchor:
\begin{equation}
\bm{A}_{v \to a} = \bm{A}^a + FF(An\text{-}CAttn_{v \to a}(\bm{Q}^a,\bm{K}^{v})\bm{A}^{v}),
\end{equation}
where $\bm{A}_{v \to a },\bm{A}_{a \to v} \in \mathbb{R}^{k \times d_t}$ denotes interaction enhanced anchor embeddings. Then we use a temporal cross attention ($T\text{-}CAttn$) by applying $\bm{E}_{a\text{-}{fused}} , \bm{E}_{v\text{-}{fused}} $ serves as query, $\bm{A}_{v \to a}$ or $\bm{A}_{a \to v}$ serves as value and key to fuse primary information and restore sequence. We get mixed embeddings note as $\bm{E}_{am}, \bm{E}_{vm} \in \mathbb{R}^{l_t \times d_t}$:
\begin{equation}
\bm{E}_{a m} = T\text{-}CAttn(\bm{A}_{v \to a},\bm{E}_{a\text{-}{fused}}),
\end{equation}
\begin{equation}
\bm{E}_{vm} = T\text{-}CAttn(\bm{A}_{a \to v},\bm{E}_{v\text{-}{fused}}),
\end{equation}
After we obtain the embedding of the modality, respectively, we add them together and feed into LayerNorm and Dropout layer to obtain the final multimodal embedding $\bm{E}_{m} \in \mathbb{R}^{l_t \times d_t}$.
\begin{equation}
\bm{E}_{m} = Dropout(LayerNorm(\bm{E}_{vm} + \bm{E}_{am})),
\end{equation}
\noindent\textbf{Multimodal Encoder.} After obtaining the multimodal embedding, we employ a pre-trained BERT\cite{devlin2019bertpretrainingdeepbidirectional} encoder to ensure the stability of textual semantics and get final embedding $\bm{E}_{f} \in \mathbb{R}^{l_t \times d_t}$ as:
\begin{equation}
\bm{E}_{f} = \text{Multimodal-Encoder}(\bm{E}_{m})
\end{equation}
\subsection{Semantic Synchronization \& Classification}
\textbf{Semantic Synchronization} To better leverage the powerful representation capabilities of large language models and enhance the representation capabilities of the \textbf{(\textit{A-ME})} module, we designed a pipeline for semantic synchronization with the representations of the LLM. Specifically, we provide the LLM with simple prompts \cite{marvin2023prompt} to automatically generate three different explanations for each label, as shown in Fig. \ref{fig:framework} (bottom). Subsequently, these explanations are forward into the pretrained SentenceBERT\cite{reimers2019sentence} to generate $\bm{S}_l=[S_1,S_2,S_3] \in \mathbb{R}^{3 \times d_t }$. $\bm{E}_{f}$ is projected into a single token $\bm{T}_{f} \in \mathbb{R}^{1 \times d_t}$ by normalizing the dimension of the sequence and the two-layer MLP. We introduce triplet contrastive learning by maximizing the similarity of negative samples descriptions and minimizing the similarity of positive content descriptions, the triplet contrastive loss is computed by:
\begin{equation}
\mathcal{L}_{tri\text{-}con} = -\frac{1}{N}\sum_{l=1}^N \log \frac{\sum_{j=1}^3\exp(sim(\bm{T}_{{f}_l}, S^+_j)/\tau)}{\sum_{k=1, k \neq i}^{N} \sum_{j=1}^3\exp(sim(\bm{T}_{{f}_l}, S^k_j)/\tau)},
\end{equation}
where $S^+$ denotes the positive samples, $S^k$ is negative samples, N represents the batch size, $sim(\cdot,\cdot)$ denotes the cosine similarity between two tokens, and $\tau$ denotes the temperature hyperparameter.\par
\textbf{Classification}
Similarly to other classification methods, we concatenate the representation embeddings with the token of semantically synchronization and feed them into a classifier, using standard cross-entropy loss to optimize the network: 
\begin{equation}
\mathcal{L}_{cls} = -\frac{1}{N}\sum_{i=1}^N\log \frac{exp(\phi([\bm{E}_{f},\bm{T}_{f}])_{label_i})}{\sum_{j=1}^L exp(\phi([\bm{E}_{f},\bm{T}_{f}])_j)},
\end{equation}
where $\phi$ is the classifier with normalization and linear layer. $label_i$ denotes the label of the sample $i^{th}$, and $L$ is the number of labels. Ultimately, The overall learning process of \textbf{(\textit{A-MESS})} is accomplished by minimizing the following loss:
\begin{equation}
    \mathcal{L} = \mathcal{L}_{tri\text{-}con} + \mathcal{L}_{cls}
\end{equation}
\subsection{Implementation Details} \label{sec:id}
To the implementation of \textbf{(\textit{A-MESS})} framework, we apply zero-padding with a maximum sequence length of 50, 180, and 400 for text, video, and audio embeddings, corresponding to $l_{t},l_{v}$ and $l_{a}$. We align the three modalities to the same length of 50 as $l_{text}$, and we unify the dimensions of the three modalities to 1024 as $d_t$ in cross attention. The experiments are conducted under PyTorch 1.13.1 \cite{paszke2019pytorch} and with an NVIDIA 3090 (24GB) GPU and CUDA 11.7, the training batch size is set to 8, the epoch is set to 40 with 8 patient epochs for early stopping as \cite{zhang2024mintrec}, $\tau$ is set to 0.7 as \cite{sun2024contextual}, the anchor selection number of the downsample $k$ is set to 8, and using AdamW \cite{loshchilov2017fixing} with a learning rate of 2e-5 to optimize the learning process. We use OpenAI GPT-4 \cite{achiam2023gpt} as explanation generator and with a ``free lunch'' prompt: \textit{You are language assistant. For the nouns I provide, please give three different descriptions. These definitions will be used for intent detection tasks. Ensure the accuracy and conciseness of the explanations, and do not output any other content. Example: Complain: 1. Expressing dissatisfaction or annoyance about something. 2. Seeking redress or resolution by voicing grievances. 3. Highlighting perceived flaws or problems in a situation. }

\section{Experiments} \label{sec:Experiments}

\begin{table*}[t]
\centering
\caption{Performance comparison of different methods on MIntRec \cite{zhang2022mintrec} and MIntRec2.0 \cite{zhang2024mintrec} in-scopes datasets where \textbf{bold} indicates the best performance, while underlined indicates the second-best performance.}
\label{tab:performance}
\begin{tabular}{l|cccc|cccc}
\toprule
\multirow{2}{*}{Methods} & \multicolumn{4}{c|}{\textbf{MIntRec\cite{zhang2022mintrec}} (ACM MM 2022)} & \multicolumn{4}{c}{\textbf{MIntRec2.0\cite{zhang2024mintrec}} (ICLR 2024)} \\
\cmidrule(lr){2-5} \cmidrule(lr){6-9}
 & ACC (\%) $\uparrow$ & F1  (\%) $\uparrow$ & P (\%) $\uparrow$ & R (\%) $\uparrow$ & ACC (\%) $\uparrow$ & F1 (\%) $\uparrow$ & P (\%) $\uparrow$ &  R (\%) $\uparrow$ \\
\midrule
\textbf{MulT}\cite{tsai2019multimodal} (ACL 2019) & 72.52 & 69.25 & 70.25 & 69.24 & 60.66 & 54.12 & 58.02 & 53.77 \\
\textbf{MAG-BERT}\cite{rahman2020integrating} (ACL 2020) & 72.65 & 68.64 & 69.08 & 69.28 & 60.58 & 55.17 & 57.78 & \underline{55.10} \\
\textbf{TCL-MAP}\cite{zhou2024token} (AAAI 2024) &  \underline{73.54} & 69.48 & 71.09 & \underline{70.27} & \underline{61.97} & \textbf{56.09} & \underline{58.14} & 53.42 \\
\textbf{CAGC}\cite{sun2024contextual} (CVPR 2024) & 73.39 & \underline{70.09} & \underline{71.21} & \textbf{70.39} & $/$ & $/$ & $/$ & $/$ \\
\midrule
\textbf{A-MESS}(Ours) & \textbf{74.12} & \textbf{70.49} & \textbf{72.95} & 69.94 & \textbf{62.39} & \underline{55.91} & \textbf{60.10} & \textbf{55.93} \\
\bottomrule
\end{tabular}
\label{tab:exp1}
\end{table*}

\subsection{Datasets}
We conduct experiments on two challenging MIR datasets to evaluate our proposed \textbf{(\textit{A-MESS})} framework. MIntRec \cite{zhang2022mintrec} is a fine-grained dataset for MIR, comprising 2,224 high-quality samples in 20 intent categories, integrating text, video and audio modalities. The dataset is divided into 1,334 training samples, 445 validation samples, and 445 test samples, with our experiments adhering to this partition. In the result of the MintRec dataset, same as \cite{zhang2022mintrec} we use the corresponding pre-trained Faster-RCNN\cite{ren2016faster} and wav2vec2.0\cite{baevski2020wav2vec20frameworkselfsupervised} as video and audio feature extractor. The subsequently released MIntRec2.0 \cite{zhang2024mintrec} is a large-scale multimodal benchmark dataset aimed at recognizing intent in conversations and detecting out-of-scope content. Compared to MIntRec, it expands to 15,000 samples, covering 30 intent categories, including approximately 9,300 in-scope and 5,700 out-of-scope annotated sentences, spanning text, video, and audio modalities. We follow the setting \cite{zhou2024token,sun2024contextual,zhang2022mintrec,zhang2024mintrec} to report the results.
\subsection{Baselines}
Following \cite{zhou2024token}, we use the following state-of-the-art MIR methods as the baselines: (1) MAG-BERT \cite{rahman2020integrating} integrates non-verbal information into pre-trained language models through the use of an MA gate. (2) MulT \cite{tsai2019multimodal} employs directional pairwise cross-modality attention to handle interactions between multimodal sequences without requiring explicit alignment. (3) TCL-MAP \cite{zhou2024token} proposes an MIR framework based on token-level contrastive learning. (4) CAGC \cite{sun2024contextual} captures rich global contextual features by mining contextual interactions within and across videos.
\subsection{Evaluation Metrics}
We assess the performance of the model using accuracy (ACC), F1 score (F1), precision (P) and recall (R), which are commonly used in classification tasks and previous work\cite{zhang2024mintrec}. Higher values in all these metrics indicate an overall improvement in performance.

\subsection{Comparison to State-of-the-Art}
\subsubsection{Results on MIntRec Dataset}
We compare \textbf{(\textit{A-MESS})} with the state-of-the-art MIR methods \cite{tsai2019multimodal,rahman2020integrating,zhou2024token,sun2024contextual} in MIntRec \cite{zhang2022mintrec}, as illustrated in Tab. \ref{tab:exp1} left. Compared to existing MIR methods, our model achieves superior performance in most metrics. It should be noted that \textbf{(\textit{A-MESS})} shows a significant improvement of 0.58\% (74.12\% vs. 73.54\%) in ACC comparing TCL-MAP, 0.4\% (70.49\% vs. 70.21\%) in the F1 score, as well as 1.74\% (72.95\% vs. 71.21\%) in precision over CAGC. Compared to state-of-the-art methods that rely solely on original labels for supervised or contrastive learning, our proposed semantic synchronization method is able to learn precise representations more efficiently.
\subsubsection{Results on MIntRec2.0 Dataset}

\begin{table}[t]
\centering
\caption{Performance comparison of baseline methods on mintrec2.0 \cite{zhang2024mintrec} in-scopes (IS) and out-of-scopes (OS) datasets.}
\label{tab:ablation}
\small 
\begin{tabular}{@{}lcccc@{}} 
\toprule
\multirow{2}{*}{Method} & \multicolumn{4}{c@{}}{IS + OS Classification (\%)} \\ 
\cmidrule(lr){2-5}
 & ACC $\uparrow$& F1-IS$\uparrow$ & F1-OS$\uparrow$ & F1$\uparrow$ \\
\midrule
\textbf{MAG-BERT}\cite{rahman2020integrating} & 56.20 & 47.52 & 62.47 & 48.00 \\
\textbf{MulT}\cite{tsai2019multimodal} & 56.00 & 46.88 & 61.66 & 47.35 \\
\textbf{A-MESS} (Ours) & \textbf{56.81} & \textbf{49.03} & \textbf{63.42} & \textbf{49.31} \\
\bottomrule
\end{tabular}
\label{tab:tb2}
\end{table}

To further evaluate the ability of \textbf{(\textit{A-MESS})}, we report the result of MIntRec2.0 as shown in Tab.\ref{tab:exp1} right, which output categories of the model and the training data are all in scope. We also report the comparison between \textbf{(\textit{A-MESS})} and several baseline methods when both the model categories and the training data include out-of-scope (OOS) samples in MIntRec2.0. Firstly, \textbf{(\textit{A-MESS})} outperforming other methods by 0.42\% (62.39\% vs.61.97) and 1.96\% (60.10\% vs. 58.14\%) in ACC and recall, respectively. This performance improvement is due to the fact that other methods do not fully recognize the importance of the semantic space in which description information on the label resides, which is crucial for learning effective representations. Meanwhile, in terms of F1 score, \textbf{(\textit{A-MESS})} almost matches the performance of existing state-of-the-art methods. Our method specifically takes into account the existence of this semantic space and introduces semantic alignment to ensure more effective representations. Furthermore, the anchor-based ME module we designed is capable of effectively learning superior multimodal information from the semantic synchronization feedback, and we will illustrate this in the ablation studies. Furthermore, we observe that our method is more accurate in identifying out-of-scope (OOS) samples, as shown in the Tab. \ref{tab:tb2}, which can be attributed to the critical semantic synchronization strategy.
\subsection{Ablation Studies}
\begin{table}[t]
\centering

\caption{Ablation study of the proposed modules on the MIntRec\cite{zhang2022mintrec} test set.}
\label{tab:ablation}
\begin{tabular}{lcccc}
\toprule
\multirow{1}{*}{Method}
 & ACC (\%) $\uparrow$ & F1 (\%) $\uparrow$ & P(\%) $\uparrow$ & R(\%) $\uparrow$ \\
\midrule
w/o A-ME & 73.24 & 69.58 & 70.07 & 69.09 \\
w/o SS & 72.88 & 68.87 & 69.30 & 69.39 \\
w/o SS and A-ME & 71.95 & 68.34 & 68.78 & 69.08 \\
A-MESS & \textbf{74.12} & \textbf{70.49} & \textbf{72.95} & \textbf{69.94} \\

\bottomrule
\end{tabular}
\label{tb:ab}
\end{table}

We evaluate the effects of the key component of \textbf{(\textit{A-MESS})}, including A-ME and SS, as illustrated in the Tab. \ref{tb:ab}. The introduction of the A-ME module resulted in an improvement in ACC of 0.88\%, while the SS module led to an increase of 1.24\%. This phenomenon indicates that these two modules exhibit independence and both contribute significantly to efficiently uncovering multimodal embeddings. Moreover, we observed that the simultaneous incorporation of the A-ME and SS modules yields a notable improvement of 4.17\% in precision. This suggests that the collaboration between the two modules and the semantic synchronization learning approach can effectively boost the representation capabilities of multimodal embeddings throughout the entire pipeline. See more details in \textbf{APPENDIX}\ref{app}.
\subsection{Anchor Analysis}
\begin{figure}[!t]
    \centering
    \begin{subfigure}[b]{0.49\linewidth}
        \includegraphics[width=\textwidth]{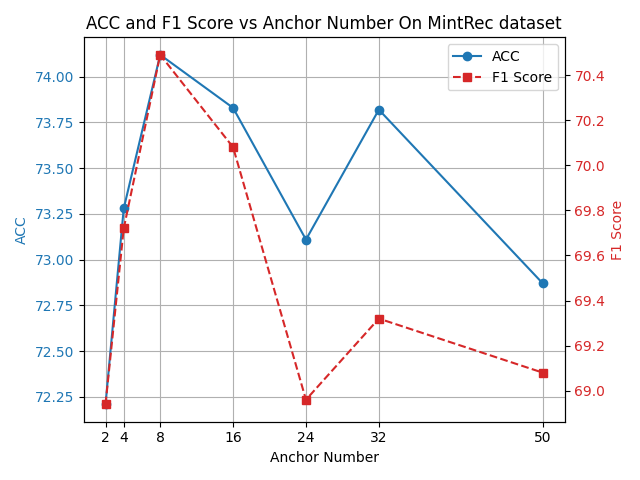}
        \label{fig:1a}
    \end{subfigure}
    \hfill
    \begin{subfigure}[b]{0.49\linewidth}
        \includegraphics[width=\textwidth]{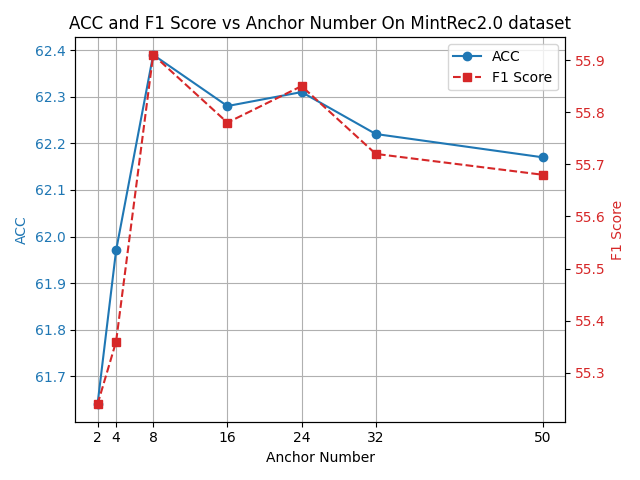}
        \label{fig:1b}
    \end{subfigure}
    \caption{Analysis of anchor number performance in MintRec\cite{zhang2022mintrec} (left) and MintRec 2.0\cite{zhang2024mintrec} (right), where \textcolor{mypaleblue}{blue} lines represent ACC, and \textcolor{red}{red} lines represent F1 score. When the number of anchors is 50, it means that no anchor is selected and all multimodal tokens are used.}
    \label{fig:an}
\end{figure}

We conduct analytical experiment on the values of k, as shown in Figure. \ref{fig:an}, it can be observed that when the number of anchors is set to 8, optimal performance is achieved independently on both datasets. Performance of the algorithm decreases when the number of anchors is either reduced or increased. This fully corroborates our original hypothesis: In auxiliary audio and video modalities, only a portion of the information can aid in multimodal representation; redundant or insufficient information can affect this representation, thus impacting the final recognition outcome.
\section{Conclusion}
In this paper, we introduce a novel approach named Anchor-based Multimodal Embedding with Semantic Synchronization \textbf{(\textit{A-MESS})} for Multimodal Intent Recognition. This method enhances the correlation between the auxiliary modalities to form efficient embeddings and performs a joint representation with the textual modality. Additionally, our framework incorporates a semantic synchronization learning method to optimize the entire framework guided by the interpretative semantics of labels produced by Large Language Models (LLMs), employing the proposed triplet contrastive learning strategy. This further enhances the representational capacity of multimodal embeddings. Experimental evaluations on two recently established benchmark datasets demonstrate that our approach achieves state-of-the-art performance and offers significant insights into multimodal learning and representation.

\bibliographystyle{IEEEbib}
\bibliography{references}

\begin{thebibliography}{10}

\bibitem{wu2020tod}
Chien-Sheng Wu, Steven Hoi, Richard Socher, and Caiming Xiong,
\newblock ``Tod-bert: Pre-trained natural language understanding for task-oriented dialogue,''
\newblock {\em arXiv preprint arXiv:2004.06871}, 2020.

\bibitem{durante2024agentaisurveyinghorizons}
Zane Durante, Qiuyuan Huang, Naoki Wake, Ran Gong, Jae~Sung Park, Bidipta Sarkar, Rohan Taori, Yusuke Noda, Demetri Terzopoulos, Yejin Choi, Katsushi Ikeuchi, Hoi Vo, Li~Fei-Fei, and Jianfeng Gao,
\newblock ``Agent ai: Surveying the horizons of multimodal interaction,'' 2024.

\bibitem{devlin2019bertpretrainingdeepbidirectional}
Jacob Devlin, Ming-Wei Chang, Kenton Lee, and Kristina Toutanova,
\newblock ``Bert: Pre-training of deep bidirectional transformers for language understanding,'' 2019.

\bibitem{zhang2022mintrec}
Hanlei Zhang, Hua Xu, Xin Wang, Qianrui Zhou, Shaojie Zhao, and Jiayan Teng,
\newblock ``Mintrec: A new dataset for multimodal intent recognition,''
\newblock in {\em Proceedings of the 30th ACM International Conference on Multimedia}, 2022, pp. 1688--1697.

\bibitem{zhang2024mintrec}
Hanlei Zhang, Xin Wang, Hua Xu, Qianrui Zhou, Kai Gao, Jianhua Su, jinyue Zhao, Wenrui Li, and Yanting Chen,
\newblock ``{MI}ntrec2.0: A large-scale benchmark dataset for multimodal intent recognition and out-of-scope detection in conversations,''
\newblock in {\em The Twelfth International Conference on Learning Representations}, 2024.

\bibitem{sun2024contextual}
Kaili Sun, Zhiwen Xie, Mang Ye, and Huyin Zhang,
\newblock ``Contextual augmented global contrast for multimodal intent recognition,''
\newblock in {\em Proceedings of the IEEE/CVF CVPR}, 2024, pp. 26963--26973.

\bibitem{zhou2024token}
Qianrui Zhou, Hua Xu, Hao Li, Hanlei Zhang, Xiaohan Zhang, Yifan Wang, and Kai Gao,
\newblock ``Token-level contrastive learning with modality-aware prompting for multimodal intent recognition,''
\newblock in {\em Proceedings of the AAAI Conference on Artificial Intelligence}, 2024, vol.~38, pp. 17114--17122.

\bibitem{rahman2020integrating}
Wasifur Rahman, Md~Kamrul Hasan, Sangwu Lee, Amir Zadeh, Chengfeng Mao, Louis-Philippe Morency, and Ehsan Hoque,
\newblock ``Integrating multimodal information in large pretrained transformers,''
\newblock in {\em Proceedings of the conference. Association for Computational Linguistics. Meeting}. NIH Public Access, 2020, vol. 2020, p. 2359.

\bibitem{tsai2019multimodal}
Yao-Hung~Hubert Tsai, Shaojie Bai, Paul~Pu Liang, J~Zico Kolter, Louis-Philippe Morency, and Ruslan Salakhutdinov,
\newblock ``Multimodal transformer for unaligned multimodal language sequences,''
\newblock in {\em Proceedings of the conference. Association for computational linguistics. Meeting}. NIH Public Access, 2019, vol. 2019, p. 6558.

\bibitem{achiam2023gpt}
Josh Achiam, Steven Adler, Sandhini Agarwal, Lama Ahmad, Ilge Akkaya, Florencia~Leoni Aleman, Diogo Almeida, Janko Altenschmidt, Sam Altman, Shyamal Anadkat, et~al.,
\newblock ``Gpt-4 technical report,''
\newblock {\em arXiv preprint arXiv:2303.08774}, 2023.

\bibitem{ezzameli2023emotion}
Kaouther Ezzameli and Hela Mahersia,
\newblock ``Emotion recognition from unimodal to multimodal analysis: A review,''
\newblock {\em Information Fusion}, vol. 99, pp. 101847, 2023.

\bibitem{zhang2020multimodal}
Chao Zhang, Zichao Yang, Xiaodong He, and Li~Deng,
\newblock ``Multimodal intelligence: Representation learning, information fusion, and applications,''
\newblock {\em IEEE Journal of Selected Topics in Signal Processing}, vol. 14, no. 3, pp. 478--493, 2020.

\bibitem{xu2023multimodal}
Peng Xu, Xiatian Zhu, and David~A Clifton,
\newblock ``Multimodal learning with transformers: A survey,''
\newblock {\em IEEE Transactions on Pattern Analysis and Machine Intelligence}, vol. 45, no. 10, pp. 12113--12132, 2023.

\bibitem{le2023multi}
Hoai-Duy Le, Guee-Sang Lee, Soo-Hyung Kim, Seungwon Kim, and Hyung-Jeong Yang,
\newblock ``Multi-label multimodal emotion recognition with transformer-based fusion and emotion-level representation learning,''
\newblock {\em IEEE Access}, vol. 11, pp. 14742--14751, 2023.

\bibitem{radford2021learningtransferablevisualmodels}
Alec Radford, Jong~Wook Kim, Chris Hallacy, Aditya Ramesh, Gabriel Goh, Sandhini Agarwal, Girish Sastry, Amanda Askell, Pamela Mishkin, Jack Clark, Gretchen Krueger, and Ilya Sutskever,
\newblock ``Learning transferable visual models from natural language supervision,'' 2021.

\bibitem{chen2020simpleframeworkcontrastivelearning}
Ting Chen, Simon Kornblith, Mohammad Norouzi, and Geoffrey Hinton,
\newblock ``A simple framework for contrastive learning of visual representations,'' 2020.

\bibitem{zhang2022dino}
Hao Zhang, Feng Li, Shilong Liu, Lei Zhang, Hang Su, Jun Zhu, Lionel~M Ni, and Heung-Yeung Shum,
\newblock ``Dino: Detr with improved denoising anchor boxes for end-to-end object detection,''
\newblock {\em arXiv preprint arXiv:2203.03605}, 2022.

\bibitem{chen2022wavlm}
Sanyuan Chen, Chengyi Wang, Zhengyang Chen, Yu~Wu, Shujie Liu, Zhuo Chen, Jinyu Li, Naoyuki Kanda, Takuya Yoshioka, Xiong Xiao, et~al.,
\newblock ``Wavlm: Large-scale self-supervised pre-training for full stack speech processing,''
\newblock {\em IEEE Journal of Selected Topics in Signal Processing}, vol. 16, no. 6, pp. 1505--1518, 2022.

\bibitem{lowe2004distinctive}
David~G Lowe,
\newblock ``Distinctive image features from scale-invariant keypoints,''
\newblock {\em International journal of computer vision}, vol. 60, pp. 91--110, 2004.

\bibitem{ZHANG2025107399}
Wenxin Zhang and Cuicui Luo,
\newblock ``Decomposition-based multi-scale transformer framework for time series anomaly detection,''
\newblock {\em Neural Networks}, vol. 187, pp. 107399, 2025.

\bibitem{vaswani2017attention}
A~Vaswani,
\newblock ``Attention is all you need,''
\newblock {\em Advances in Neural Information Processing Systems}, 2017.

\bibitem{marvin2023prompt}
Ggaliwango Marvin, Nakayiza Hellen, Daudi Jjingo, and Joyce Nakatumba-Nabende,
\newblock ``Prompt engineering in large language models,''
\newblock in {\em International conference on data intelligence and cognitive informatics}. Springer, 2023, pp. 387--402.

\bibitem{reimers2019sentence}
N~Reimers,
\newblock ``Sentence-bert: Sentence embeddings using siamese bert-networks,''
\newblock {\em arXiv preprint arXiv:1908.10084}, 2019.

\bibitem{paszke2019pytorch}
Adam Paszke, Sam Gross, Francisco Massa, Adam Lerer, James Bradbury, Gregory Chanan, Trevor Killeen, Zeming Lin, Natalia Gimelshein, Luca Antiga, et~al.,
\newblock ``Pytorch: An imperative style, high-performance deep learning library,''
\newblock {\em Advances in neural information processing systems}, vol. 32, 2019.

\bibitem{loshchilov2017fixing}
Ilya Loshchilov, Frank Hutter, et~al.,
\newblock ``Fixing weight decay regularization in adam,''
\newblock {\em arXiv preprint arXiv:1711.05101}, vol. 5, 2017.

\bibitem{ren2016faster}
Shaoqing Ren, Kaiming He, Ross Girshick, and Jian Sun,
\newblock ``Faster r-cnn: Towards real-time object detection with region proposal networks,''
\newblock {\em IEEE transactions on pattern analysis and machine intelligence}, vol. 39, no. 6, pp. 1137--1149, 2016.

\bibitem{baevski2020wav2vec20frameworkselfsupervised}
Alexei Baevski, Henry Zhou, Abdelrahman Mohamed, and Michael Auli,
\newblock ``wav2vec 2.0: A framework for self-supervised learning of speech representations,'' 2020.

\end{thebibliography}

\clearpage
\onecolumn
\appendix\label{app}

\section*{\Large Result of Prompting}
In our experimental setup, we supplied prompts to OpenAI’s GPT-4, instructing it to generate the corresponding verbal descriptions, this straightforward prompt nearly represents a "free lunch" in terms of its minimal cost and significant benefit. \textit{You are language assistant. For the nouns I provide, please give three different descriptions. These definitions will be used for intent detection tasks. Ensure the accuracy and conciseness of the explanations, and do not output any other content. Example: Complaint: 1. Holding dissatisfaction in mind and blaming others. 2. Expressing discontent about something to vent emotions. 3. Feeling restless and grumbling.} \par
\textbf{The results we used are presented below:}\par
\textbf{In MintRec\cite{zhang2022mintrec}:} \par
1. **Complain**  
   1. To express dissatisfaction or annoyance about something.  
   2. To seek resolution by voicing grievances.  
   3. To highlight perceived flaws or problems in a situation.

2. **Praise**  
   1. To express approval or admiration for someone's actions or qualities.  
   2. To acknowledge and celebrate positive achievements or behavior.  
   3. To commend or compliment someone or something.

3. **Apologize**  
   1. To express regret for an action or situation that caused harm or offense.  
   2. To offer an acknowledgment of wrongdoing or mistake.  
   3. To ask for forgiveness or understanding from someone affected.

4. **Thank**  
   1. To express gratitude or appreciation for something received.  
   2. To convey recognition for someone's kindness or favor.  
   3. To acknowledge someone's help or gesture with gratitude.

5. **Criticize**  
   1. To evaluate or analyze the merits and faults of something or someone.  
   2. To express disapproval or point out flaws in an action or behavior.  
   3. To provide negative feedback or highlight areas of improvement.

6. **Care**  
   1. To show concern or interest for someone or something.  
   2. To provide attention and nurturing to maintain well-being.  
   3. To have responsibility or take action to protect or support.

7. **Agree**  
   1. To express acceptance or approval of an idea, statement, or proposal.  
   2. To share the same opinion or belief about something.  
   3. To reach a mutual decision or consensus.

8. **Taunt**  
   1. To mock or provoke someone in a disrespectful manner.  
   2. To ridicule or tease someone to provoke a reaction.  
   3. To make insulting or sarcastic remarks toward someone.

9. **Flaunt**  
   1. To display something conspicuously, often to impress others.  
   2. To show off or boast about an achievement or possession.  
   3. To make a deliberate display of wealth, success, or ability.

10. **Oppose**  
    1. To resist or act against something or someone.  
    2. To disagree with a particular idea, proposal, or action.  
    3. To stand in contrast or opposition to a position or belief.

11. **Joke**  
    1. A humorous statement or anecdote intended to entertain or amuse.  
    2. A playful remark made to provoke laughter or lighten the mood.  
    3. A form of humor that involves wit or absurdity to elicit a reaction.

12. **Inform**  
    1. To provide someone with knowledge or facts about a topic.  
    2. To give updates or details on a situation.  
    3. To notify or make someone aware of important information.

13. **Advise**  
    1. To offer suggestions or recommendations based on knowledge or experience.  
    2. To guide or counsel someone on a course of action.  
    3. To provide expert or informed opinions to help make decisions.

14. **Arrange**  
    1. To organize or put things into a specific order or structure.  
    2. To make preparations or plans for an event or activity.  
    3. To coordinate the details of a meeting, event, or task.

15. **Introduce**  
    1. To present someone to others, typically for the first time.  
    2. To bring attention to a new concept, idea, or product.  
    3. To make someone or something known in a particular context.

16. **Comfort**  
    1. To provide emotional support to alleviate distress or sadness.  
    2. To offer reassurance and encouragement in difficult times.  
    3. To console or ease the discomfort of someone experiencing hardship.

17. **Leave**  
    1. To depart from a location or situation.  
    2. To give up possession or control of something.  
    3. To discontinue participation or involvement in an activity.

18. **Prevent**  
    1. To stop something from happening or arising.  
    2. To take measures to avoid potential risks or problems.  
    3. To hinder or restrict the occurrence of an event or situation.

19. **Greet**  
    1. To acknowledge or welcome someone upon arrival.  
    2. To express a polite or friendly greeting.  
    3. To address someone with a salutation, often at the beginning of an interaction.

20. **Ask for Help**  
    1. To request assistance or support from others.  
    2. To seek guidance or advice in solving a problem.
    3. To ask someone to take action or provide aid in a task. \par
\textbf{and in  MintRec2.0 \cite{zhang2024mintrec}:} \par
1. **Doubt**  
   1. A feeling of uncertainty or lack of conviction about something.  
   2. Questioning the truth or validity of a statement or belief.  
   3. Hesitation or reluctance to accept a claim or idea.

2. **Acknowledge**  
   1. To recognize or admit the existence or truth of something.  
   2. To express thanks or gratitude for something received.  
   3. To respond to or indicate awareness of a communication or action.

3. **Refuse**  
   1. To decline or reject an offer or request.  
   2. To indicate unwillingness to accept or comply with something.  
   3. To deny permission or agreement to a proposal.

4. **Warn**  
   1. To inform someone of a potential danger or risk.  
   2. To alert others about something that could cause harm or trouble.  
   3. To give a precautionary signal or advice.

5. **Emphasize**  
   1. To highlight or give special attention to something.  
   2. To stress the importance or significance of a point.  
   3. To make something more noticeable or prominent.

6. **Complain**  
   1. To express dissatisfaction or annoyance about something.  
   2. To seek resolution by voicing grievances.  
   3. To highlight perceived flaws or problems in a situation.

7. **Praise**  
   1. To express approval or admiration for someone's actions or qualities.  
   2. To acknowledge and celebrate positive achievements or behavior.  
   3. To commend or compliment someone or something.

8. **Apologize**  
   1. To express regret for an action or situation that caused harm or offense.  
   2. To offer an acknowledgment of wrongdoing or mistake.  
   3. To ask for forgiveness or understanding from someone affected.

9. **Thank**  
   1. To express gratitude or appreciation for something received.  
   2. To convey recognition for someone's kindness or favor.  
   3. To acknowledge someone's help or gesture with gratitude.

10. **Criticize**  
    1. To evaluate or analyze the merits and faults of something or someone.  
    2. To express disapproval or point out flaws in an action or behavior.  
    3. To provide negative feedback or highlight areas of improvement.

11. **Care**  
    1. To show concern or interest for someone or something.  
    2. To provide attention and nurturing to maintain well-being.  
    3. To have responsibility or take action to protect or support.

12. **Agree**  
    1. To express acceptance or approval of an idea, statement, or proposal.  
    2. To share the same opinion or belief about something.  
    3. To reach a mutual decision or consensus.

13. **Oppose**  
    1. To resist or act against something or someone.  
    2. To disagree with a particular idea, proposal, or action.  
    3. To stand in contrast or opposition to a position or belief.

14. **Taunt**  
    1. To mock or provoke someone in a disrespectful manner.  
    2. To ridicule or tease someone to provoke a reaction.  
    3. To make insulting or sarcastic remarks toward someone.

15. **Flaunt**  
    1. To display something conspicuously, often to impress others.  
    2. To show off or boast about an achievement or possession.  
    3. To make a deliberate display of wealth, success, or ability.

16. **Joke**  
    1. A humorous statement or anecdote intended to entertain or amuse.  
    2. A playful remark made to provoke laughter or lighten the mood.  
    3. A form of humor that involves wit or absurdity to elicit a reaction.

17. **Ask for Opinions**  
   1. To request someone's thoughts or views on a topic.  
   2. To seek feedback or advice on a matter.  
   3. To inquire about preferences or judgments from others.

18. **Confirm**  
   1. To verify or establish the truth of something.  
   2. To affirm or give assurance about a statement or decision.  
   3. To validate or acknowledge receipt or understanding of information.

19. **Explain**  
   1. To clarify or provide additional details about something.  
   2. To describe the reasoning or causes behind an event or situation.  
   3. To make something understandable by breaking it down or offering examples.

20. **Invite**  
   1. To extend an offer or request for someone to join an event or activity.  
   2. To ask someone to participate or engage in a particular event or gathering.  
   3. To encourage or request someone's presence or participation in something.

21. **Plan**  
   1. To organize or arrange the steps required to achieve a goal.  
   2. To make decisions about future actions or strategies.  
   3. To prepare or outline the details for an event or course of action.

22. **Inform**  
   1. To provide someone with knowledge or facts about a topic.  
   2. To give updates or details on a situation.  
   3. To notify or make someone aware of important information.

23. **Advise**  
   1. To offer suggestions or recommendations based on knowledge or experience.  
   2. To guide or counsel someone on a course of action.  
   3. To provide expert or informed opinions to help make decisions.

24. **Arrange**  
   1. To organize or put things into a specific order or structure.  
   2. To make preparations or plans for an event or activity.  
   3. To coordinate the details of a meeting, event, or task.

25. **Introduce**  
   1. To present someone to others, typically for the first time.  
   2. To bring attention to a new concept, idea, or product.  
   3. To make someone or something known in a particular context.

26. **Comfort**  
    1. To provide emotional support to alleviate distress or sadness.  
    2. To offer reassurance and encouragement in difficult times.  
    3. To console or ease the discomfort of someone experiencing hardship.

27. **Leave**  
    1. To depart from a location or situation.  
    2. To give up possession or control of something.  
    3. To discontinue participation or involvement in an activity.

28. **Prevent**  
    1. To stop something from happening or arising.  
    2. To take measures to avoid potential risks or problems.  
    3. To hinder or restrict the occurrence of an event or situation.

29. **Greet**  
    1. To acknowledge or welcome someone upon arrival.  
    2. To express a polite or friendly greeting.  
    3. To address someone with a salutation, often at the beginning of an interaction.

30. **Ask for Help**  
    1. To request assistance or support from others.  
    2. To seek guidance or advice in solving a problem.  
    3. To ask someone to take action or provide aid in a task.\par
    
UNKOWN : \textit{\textbf{For this category, we do not apply triplet comparative learning}}

\section{Prompt Quantity Analysis}
\begin{table*}[!t]
  \centering
  \caption{Sample Nums Comparing On MintRec \cite{zhang2022mintrec}}
  \label{tab:pos_neg_samples}
  \begin{tabular}{cccc}
    \toprule
     Number of descriptions & ACC & F1 \\
    \midrule
    2              & 73.83                 & 70.11                 \\
    3              & \textbf{74.12}                 &  \textbf{70.49}                 \\
    5              & 74.07                & 70.29                 \\
    8              & 73.78                 & 70.07                 \\
    \bottomrule
  \end{tabular}
\label{lb1}
\end{table*}

We evaluate the influence of pos / neg sample counts on MintRec\cite{zhang2022mintrec}, i.e., the amount of descriptions generated, on the A-MESS method during contrastive learning within the semantic synchronization module. We found that generating only three descriptions was sufficient to achieve optimal performance, which is present in \textbf{bold}.\par
We attribute this phenomenon to insight into representation learning: In the representation space (considered as a two-dimensional plane for simplicity), while two points can define a line segment, the inherent instability of the representation learning process causes multimodal representations to fluctuate around this line, making it challenging to achieve perfect fitting and complicating algorithm convergence. However, based on geometric principles, at least three points can define a subplane in the representation space. According to the idea of contrastive learning, within this subplane, a representation is considered correct if it falls within this plane. This approach not only significantly stabilizes the algorithm's performance, but also accelerates convergence and aligns better with human intuition.

Furthermore, from the data presented in Table\ref{lb1}, we observe that increasing the number of representations for contrastive learning does not enhance performance and may even lead to a slight decline. This observation further validates our insight that constraining learning within a smaller, more stable representation sub-plane yields superior results.

\section*{\Large Details of Anchor selection}
To more clearly elucidate the proposed A-ME module, the implementation details of the anchor selection component are presented here. For the given $\bm{E}_{text} \in \mathbb{R}^{l_t \times d_t}$ and $\{\bm{E}_{\alpha\text{-}fused},\bm{E}_{v\text{-}fused} \} \in \mathbb{R}^{l_t \times d_t }$, pairwise similarities are computed to obtain $\{\bm{A}^{\alpha},\bm{A}^v \} \in \mathbb{R}^{k \times d_t }$. The ratios between the nearest and the second nearest neighbors are then ranked to determine the confidence scores, from which the top-k indices are selected. The corresponding tokens in the original embeddings are chosen as auxiliary anchors A according to these indices. The pseudocode is presented as follows:
\begin{center}
    \begin{lstlisting}[language=Python,caption=$AnchorSelection-Pseudo Code$]

def select_anchor_tokens(t_embeddings, v_a_embeddings, top_k=8):
    batch_size, seq_len, dim = t_embeddings.shape

    source_norm = t_embeddings / t_embeddings.norm(dim=-1, keepdim=True)
    target_norm = v_a_embeddings / v_a_embeddings.norm(dim=-1, keepdim=True)
    # similarity_matrix [batch_size, seq_len, seq_len]
    similarity_matrix = torch.matmul(source_norm, target_norm.transpose(1, 2)) 
    sorted_similarities, indices = torch.sort(similarity_matrix, dim=-1, descending=True)
    first_nearest_similarity = sorted_similarities[:, :, 0] 
    second_nearest_similarity = sorted_similarities[:, :, 1] 
    _, top_k_indices = torch.topk(reliability_score, top_k, dim=-1)  
    # source_anchor_features [batch_size, top_k, dim]
    source_anchor_features = v_a_embeddings[torch.arange(batch_size).unsqueeze(-1), top_k_indices]

    return source_anchor_features

# inference:
v_T_Anchor_embedding = select_anchor_tokens(text_embedding,V_fused)
A_T_Anchor_embedding = select_anchor_tokens(text_embedding,A_fused)

    \end{lstlisting}
\end{center}
\label{code}

\section*{\Large Semantic Synchronization Analysis}\par
We evaluate the distribution of the token $\mathbf{T}_f$ both before and after applying semantic synchronization with triplet contrastive learning in MintRec \cite{zhang2022mintrec}. Specifically, we first compute the initial $\mathbf{T}_f^{mean}$ by averaging the feature $\mathbf{E}_f$ directly. Next, we obtain $\mathbf{T}_f^{SS}$ after applying semantic synchronous contrastive learning and perform principal component analysis (PCA) with the explanatory embeddings of the current category to reduce the data to a two-dimensional space. Subsequently, we normalize these reduced-dimension results for comparative analysis. We randomly selected four categories as Agree, Joke, Criticize, and Oppose for reporting. As shown in Fig. \ref{fig:semantic}.\par
We found that token representations after semantic synchronization significantly approach the semantic space of the labels. This result fully validates our approach and indicates that the synchronized semantic space can reasonably improve classification accuracy.

\begin{figure}[ht]
    \centering
    \begin{subfigure}[b]{0.49\linewidth}
        \includegraphics[width=\textwidth]{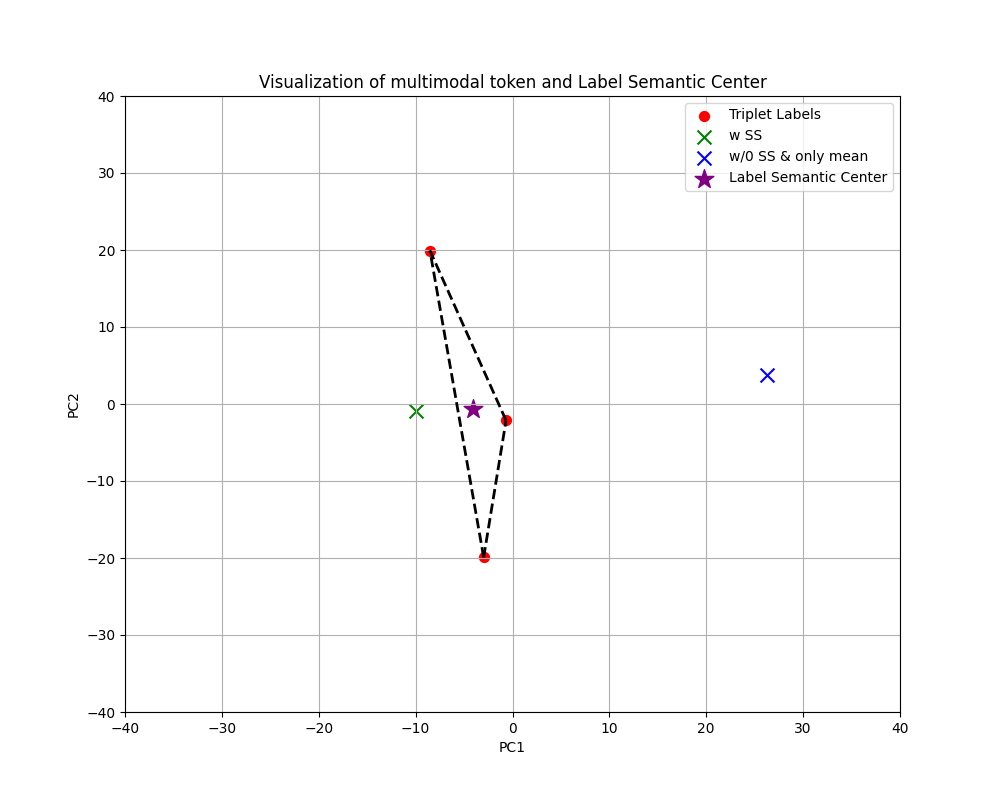}
        \caption{Agree}
    \end{subfigure}
    \hfill
    \begin{subfigure}[b]{0.49\linewidth}
        \includegraphics[width=\textwidth]{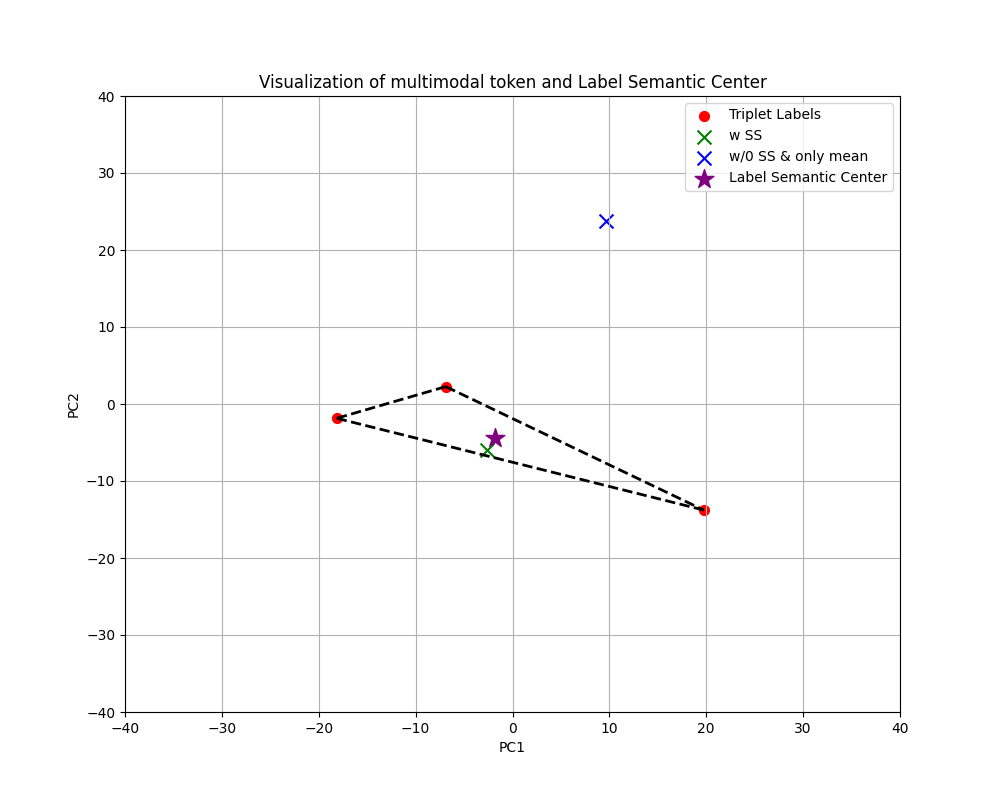}
        \caption{Joke}
    \end{subfigure}
    \hfill
        \begin{subfigure}[b]{0.49\linewidth}
        \includegraphics[width=\textwidth]{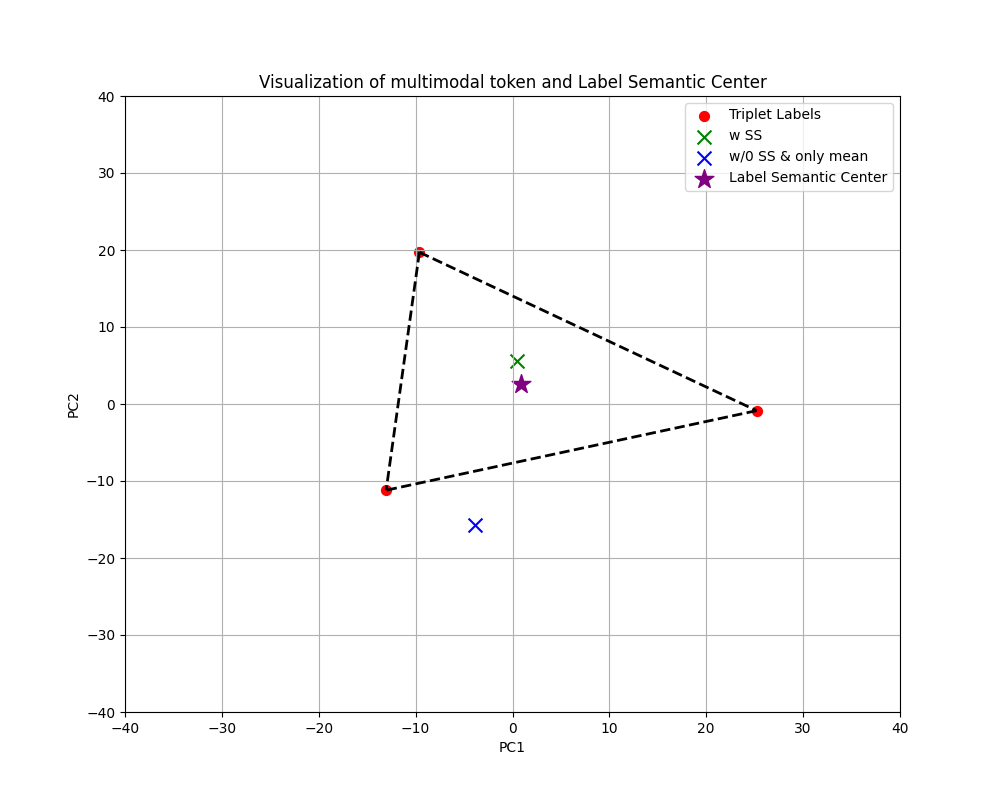}
        \caption{Criticize}
    \end{subfigure}
    \hfill
    \begin{subfigure}[b]{0.49\linewidth}
    \includegraphics[width=\textwidth]{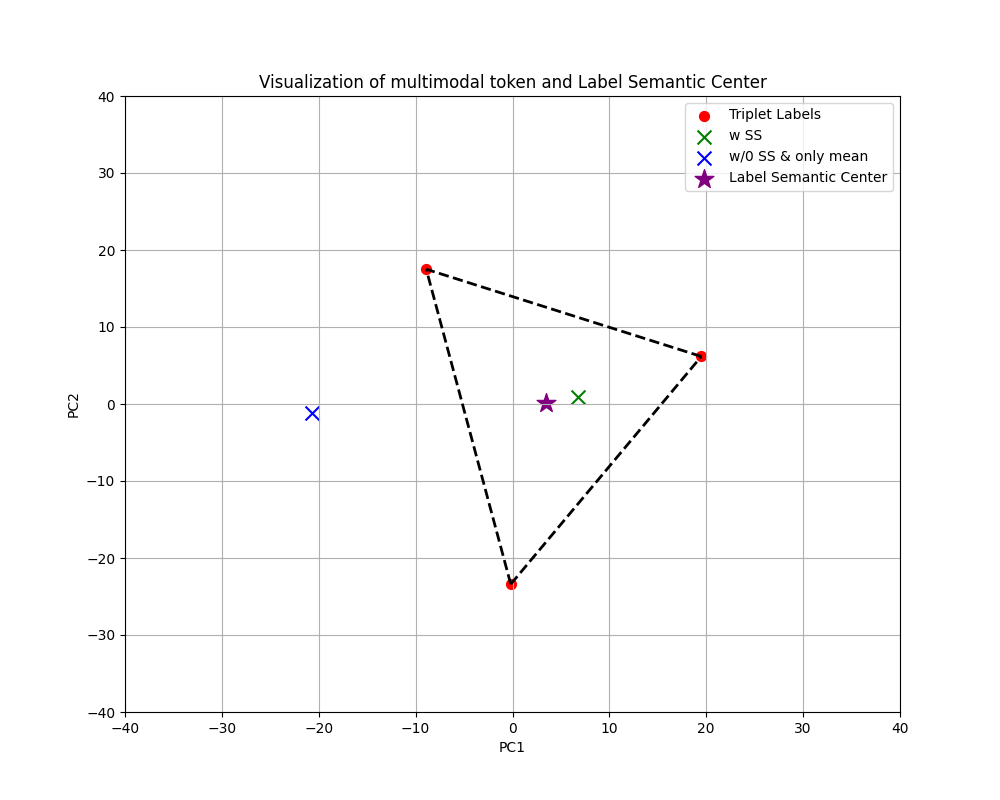}
    \caption{Oppose}
    \label{fig:1b}
    \end{subfigure}
    \caption{Analysis of semanctic, $\mathbf{T}_f^{mean}$ in \textcolor{blue}{blue} and $\mathbf{T}_f^{SS}$ in \textcolor{green}{green}, The dotted line indicates the semantic plane of the label.}
\end{figure}
\label{fig:semantic}

\end{document}